# Do LLMs Understand Why We Write Diaries?
# A Method for Purpose Extraction and Clustering

Valeriya Goloviznina[1][0000-0003-1167-2606], Alexander Sergeev[1][0009-0002-4103-842X],
Mikhail Melnichenko[1][0009-0001-2370-3193], Evgeny Kotelnikov[1][0000-0001-9745-1489]

[1] European University at Saint Petersburg, St. Petersburg, Russia
{v.goloviznina, a.sergeev, mmelnichenko, e.kotelnikov}@eu.spb.ru

**Abstract.** Diary analysis presents challenges, particularly in extracting meaningful information from large corpora, where traditional methods often fail to deliver satisfactory results. This study introduces a novel method based on Large Language Models (LLMs) to identify and cluster the various purposes of diary writing. By "purposes," we refer to the intentions behind diary writing, such as documenting life events, self-reflection, or practicing language skills. Our approach is applied to Soviet-era diaries (1922–1929) from the Prozhito digital archive, a rich collection of personal narratives. We evaluate different proprietary and open-source LLMs, finding that GPT-4o and o1-mini achieve the best performance, while a template-based baseline is significantly less effective. Additionally, we analyze the retrieved purposes based on gender, age of the authors, and the year of writing. Furthermore, we examine the types of errors made by the models, providing a deeper understanding of their limitations and potential areas for improvement in future research.

**Keywords:** LLM, Diaries Analysis, Clustering, Archival Documents, Digital Humanities.

## 1    Introduction

Large Language Models (LLMs) are widely used in Digital Humanities (DH) and Computational Social Sciences (CSS) for tasks such as analyzing archives and forums, topic modeling, cultural analytics, text classification, summarization, and generating human-like text with interpretable explanations across fields such as sociology, psychology, literature, history, and linguistics [1, 2, 3].

One of the important fields of application of LLMs is the analysis of archival documents such as diaries, letters, journals, etc. It is difficult or impossible to read huge text corpora if it is necessary to extract some important elements from the texts. Simple template-based methods do not work efficiently enough. This is where LLMs come to the rescue.

However, in practice, DH and CSS researchers face the following challenges:

1. uncertainty in model selection and in the accuracy of the model for a particular problem;

2. difficulty in using standard interfaces to LLMs to analyze large corpora;
3. the need for programming skills to access the APIs and handle the results;
4. the high cost of inference.

In this paper, we propose a method to analyze the large diary corpora. In the human sciences, the diary is used as a qualitative research tool [4]. The diary analysis allows researchers to study people's experiences, behaviors, and life circumstances in a natural setting. This research area has been actively explored from the 1990s to the present [5]. Among other things, researchers are interested in the purposes of keeping a diary [6].

In our work, we extract the purposes of diary writing using LLMs. By purposes, we mean the intentions with which people keep diaries – both the reasons why a person started writing (for example, inspiration from someone else's diary, boredom, free time) and the purposes for doing so (for example, writing memoirs, tracking your condition, practicing foreign languages).

As data we use the corpus of diaries of the post-revolutionary Soviet era (1922–1929) from "Prozhito", the Center for the Study of Ego-Documents of the European University at St. Petersburg, which has been collecting and publishing personal diaries in Russian and other languages since 2015 and actively developing a digital archive[1].

The contributions of our work are as follows:

1. we test different proprietary and open-source models for identifying diary entries that contain purposes and for extracting these purposes;
2. we evaluate the performance of these models on both tasks – identification and extraction – based on manual annotation of the results;
3. we propose an iterative algorithm that leverages LLMs to cluster the extracted purposes and evaluate the results for various LLMs;
4. in addition to advancing Computational Linguistics, our study contributes to Digital Humanities and Computational Social Sciences by proposing a scalable and accessible methodology for quantitative research on personal narratives;
5. this study contributes to the humanities by examining the diverse motivations behind diary writing. By analyzing gender, age, and temporal factors, we uncover the significance of diary entries as tools for self-expression and cultural documentation.

## 2     Previous Work

LLMs are often used to extract structured data from unstructured documents. For instance, in materials science and chemical research, LLMs are employed to extract chemical knowledge, such as formulas or bandgap values [7, 8]. In agriculture, they are applied for pest identification [9], and in medicine, they are used to extract symptoms of pulmonary embolism 10]. Baddour propose using LLMs to search for phenotypes in clinical reports related to medical research [11]. The integration of a novel LLM-based span detector component in their work improves results.

---

[1] https://prozhito.org

In addition to these applications, LLMs are utilized for diary analysis [6, 12, 13]. Diary analysis involves collecting qualitative information about individuals' daily lives and is applied in fields such as psychology, education, and healthcare to study human behavior [14].

However, processing such data is challenging, as it requires substantial time and effort. Modern technologies, including LLMs, can streamline this process by accelerating data analysis and providing deeper insights into diary content.

Pooley investigated the reasons why individuals begin and cease diary-keeping [6]. However, their study relied on labor-intensive manual analysis rather than leveraging LLMs. In contrast, Shin et al. used diary analysis to identify depression, employing the GPT-3.5 and GPT-4 models [12]. Using 428 diaries from 91 participants, the fine-tuned GPT-3.5 demonstrated higher performance in detecting depression, achieving an accuracy of 0.902 and an F1-score of 0.685.

Li et al. introduced DiaryHelper, a tool that utilizes generative AI techniques to assist diary writers in capturing event details with minimal effort, thereby reducing bias in the note-taking process [13]. DiaryHelper predicts five dimensions of information critical for episodic memory – time, place, emotion, people, and activity – for each recorded event. The GPT-3.5 model is used to generate possible labels for these dimensions.

A related application of LLMs is in social networking analysis. Alhamed et al. used the LLaMA-2-7B-Chat model to detect evidence of suicidal tendencies in Twitter posts, achieving an accuracy of 0.96 [15].

The reviewed studies demonstrate the effectiveness of LLMs in extracting data from unstructured documents and analyzing diary entries. However, none of the studies address the extraction of diary purposes. Furthermore, unlike the reviewed works, our study employs different state-of-the-art LLMs, enabling a comparative evaluation of their performance in extracting data from unstructured documents.

## 3 Methodology

Our methodology includes three steps – purpose extraction by different models, annotation and evaluation of the extracted sets, and purpose clustering.

**Step 1. Purpose Extraction.** We process the entire corpus using multiple models with the same prompt (see Appendix A). On each run, 10 text diary entries are fed (no more than 15,000 tokens[2]), from which models extract potential purposes.

**Step 2. Annotation and Evaluation of Extracted Sets.** The union of the extracted sets was given to the three annotators for labeling. Annotators had to answer the question, "Is the purpose(s) of diary writing present in this diary entry, and if so, did the model(s) correctly extract that purpose(s)?"[3] A binary answer was implied in both questions. As a result, a set of diary entries containing purposes was identified, relative to which performance scores were computed. Also, for such entries, scores of the purposes extracted by the models were obtained.

---

[2] The minimum of the limits of the used models.

[3] Annotators were shown extracted purposes only if they indicated that the entry had a purpose.

We used Precision, relative Recall, and relative F1-score as performance scores. Precision, defined as the proportion of correct answers given by the model to the total number of answers provided by the model, serves as an objective measure in this context. Recall, as the proportion of correct answers of a model to the total number of correct answers in the whole corpus, cannot be properly computed because we are not able to label the whole corpus. This limitation arises from the labor-intensive of annotating.

To achieve a more comprehensive evaluation, we compute relative Recall and relative F1-score, following the approach used in studies [16, 17], for example. In our case, we consider the set of all correct answers to be the set of correct answers labeled by the annotators.

In addition to evaluating individual models, we also assess unions of these models, hypothesizing that merging their answers may improve the performance of purpose extraction.

**Step 3. Purpose Clustering.** The number of purposes extracted by the models turns out to be quite large (hundreds of purposes), so it is necessary to combine them into categories (clusters). We evaluate how well LLMs cope with purpose clustering.

In the simplest case, the model is given all purposes as input and asked to cluster them. However, none of the models was able to successfully cluster the purposes in one run because a large number of purposes did not belong to any cluster.

Therefore, we used the following algorithm.

1. Clustering Initialization: provide the model with a complete list of purposes and ask it to generate cluster names (see the prompt in the Appendix B).
2. Purpose Assignment: provide the model with a current list of purposes (initially including all purposes) along with the generated cluster names. Ask the model to assign the purposes to the clusters (see the prompt in the Appendix C). Check which purposes have been assigned to clusters – exclude them from the list of purposes.
3. Repeat step 2 until the list of purposes is empty.

To evaluate the quality of the clustering, we manually partitioned each set of purposes into clusters and used these partitions as references. As a performance measure, we used the Rand index [18] and averaged it over all sets of purposes.

## 4    Data and Models

### 4.1    Data

The texts of 40,222 personal diaries of 247 authors written in Russian between January 1, 1922 and December 31, 1929 were used as the corpus. Entries that were too short (consisting of 1–2 words) and too long (exceeding 1,400 tokens for any of the tested models) were removed from this set. The value of 1,400 was derived from the context limitations of the LLM – the prompt included 10 entries, the lower bound on the context limit was 15,000 tokens (DeepSeek V3 model, provided by DeepInfra[4], at the time of the experiments). This leaves 38,332 diary entries.

Distributions of diary entries gender, author age category, and period of entries writing presented in Table 1–3.

The corpus is unbalanced by gender. Experts offer the following explanations for this imbalance: women destroyed their diaries; women kept fewer diaries due to the large number of daily responsibilities, including those related to childbirth.

The time period of entries writing from 1922 to 1929 is divided into three parts: the early years of the decade (1922–1923), the middle years (1924–1926), and the late years (1927–1929).

To determine age categories, we rely on Levinson's stages: pre-adulthood (under 17 years old), early adulthood (ages 18-39), middle adulthood (ages 40-59), and late adulthood (60 years old and older) [19].

**Table 1**. Distribution of entries by gender.

| Gender | # entries |
| --- | --- |
| Male | 32,371 |
| Female | 5,961 |

**Table 2**. Distribution of entries by author age category.

| Age category | # entries |
| --- | --- |
| under 17 years old | 2,514 |
| 18-39 | 14,381 |
| 40-59 | 15,323 |
| over 60 years old | 5,287 |

**Table 3**. Distribution of entries by period of entries writing.

| Period of writing | # entries |
| --- | --- |
| early 1920s (1922–1923) | 8,302 |
| mid 1920s (1924–1926) | 14,021 |
| late 1920s (1927–1929) | 16,009 |

### 4.2 Models

The following models were considered:

- **OpenAI GPT-4o** (v. 2024-08-06) – a proprietary chat model;
- **OpenAI o1-mini** (v. 2024-09-12) – a proprietary reasoning model;
- **DeepSeek-V3** – an open-source chat model with MoE architecture, has 671B total parameters and 37B active parameters to generate each token [20]. Hereafter, we will refer to DeepSeek-V3 simply as DeepSeek.

---

[4] https://deepinfra.com

For the OpenAI models their own provider was used, for the other models the Deep-Infra provider was used.

Statistics of the corpus processing for each model are presented in Table 4.

**Table 4.** Statistics of the corpus processing for each model. Number and average number of tokens are the parameters of the corpus with respect to model tokenizers. The number of entries and purposes shows for how many diary entries the model indicated the number of purposes and how many total purposes the model extracted. The cost includes processing of the entire corpus and clustering of purposes (for OpenAI – using BatchAPI).

| Model | # tokens | Avg. tokens | # entries | # purposes | Processing cost |
|---|---|---|---|---|---|
| baseline[5] | 3,868,282 | 100.9±114.4 | 177 | 177 | $0 |
| GPT-4o | 8,111,474 | 211.6±226.6 | 108 | 172 | $9.81 |
| o1-mini | 8,111,474 | 211.6±226.6 | 71 | 118 | $16.70 |
| DeepSeek | 9,183,438 | 239.6±256.3 | 290 | 443 | $5.81 |

We also tried to use other models – OpenAI GPT-4o-mini, DeepSeek-R1-Distill-Llama-70B, Qwen-2.5-72B-Instruct – but the number of entries in which they found purposes was too large (GPT-4o-mini – 2,983, DeepSeek-R1 – 685, Qwen-2.5 – 13,405) and preliminary analysis showed a large number of incorrectly selected entries, so it was decided not to analyze them in more depth.

Besides LLMs, we tested a simple template-based model (baseline). We created lists of nouns ("diary", "record", "purpose", etc.) and verbs ("keep", "write", "help", etc.) indicating the purpose of keeping a diary. An entry was considered to potentially contain a purpose if any of the possible morphology-aware noun+verb pairs occurred in it. In this case, the purpose was defined as the sentence in the entry that contained this pair. Using the baseline, we identified 177 entries, corresponding to 177 purposes.

## 5    Results

**Step 1. Purpose Extraction.** We obtained results from 4 models for 38,332 diary entries. The number of entries for which the models extracted purposes is shown in Table 5. The union of the four sets includes 460 entries. These entries were given to three annotators for labeling.

**Table 5.** Results of the models (first four rows) and the union of the models (last four rows). *Entries* is the number of diary entries in which the model identified purposes, *Correct entries* is the number of diary entries labeled by annotators as actually containing purposes.

| Model | # entries | # correct entries | Precision | Rel. Recall | Rel. F1-score |
|---|---|---|---|---|---|
| baseline | 177 | 35 | 0.1977 | 0.2059 | 0.2017 |
| GPT-4o | 108 | 83 | **0.7685** | 0.4882 | 0.5971 |
| o1-mini | 71 | 48 | 0.6761 | 0.2824 | 0.3983 |
| DeepSeek | 290 | 140 | 0.4828 | 0.8235 | 0.6087 |

---

[5] In baseline, tokens are words.

Continuation of Table 5

| Model | # entries | # correct entries | Precision | Rel. Recall | Rel. F1-score |
|---|---|---|---|---|---|
| GPT-4o ∪ o1-mini | 153 | 107 | 0.6993 | 0.6294 | **0.6625** |
| GPT-4o ∪ DeepSeek | 306 | 147 | 0.4804 | 0.8647 | 0.6177 |
| o1-mini ∪ DeepSeek | 319 | 156 | 0.4890 | 0.9177 | 0.6380 |
| GPT-4o ∪ o1-mini ∪ DeepSeek | 333 | 161 | 0.4834 | **0.9471** | 0.6402 |

**Step 2. Annotation and Evaluation of Extracted Sets.** From 460 diary entries, the annotators identified 170 entries by majority voting as actually containing the purpose(s). The performance scores of the individual models, as well as their union[6], are shown in the Table 5. The inter-annotator agreement was quite high: Krippendorff's alpha=0.895.

The best result for Precision is shown by GPT-4o, o1-mini lags behind by 10 p.p. DeepSeek extracts a lot of entries (290), but only 140 of them are correct, which results in low Precision. However, this is compensated by high relative Recall, so relative F1-score is even slightly higher than GPT-4o. Among the model union, the best result is the union of GPT-4o and o1-mini in terms of Precision and relative F1-score. This union represents the optimal balance of Precision and relative Recall.

For diary entries that annotators labeled as containing purposes (170 entries), the purposes extracted by the models were also labeled. The results are shown in Table 6. GPT-4o and o1-mini show comparable results, with DeepSeek lagging behind by 5–6 pp.

**Table 6**. Results of the purpose extraction. *Purposes* refers to the number of purposes extracted by the model, considering only those diary entries that annotators labeled as containing a purpose. *Mean* is the average number of extracted purposes per entry. *Correct purposes* is the number of purposes labeled by annotators as valid purposes.

| Model | # purposes | mean | # correct purposes | Precision |
|---|---|---|---|---|
| GPT-4o | 141 | 1.70 | 126 | 0.8936 |
| o1-mini | 85 | 1.77 | 77 | **0.9059** |
| DeepSeek | 221 | 1.58 | 187 | 0.8462 |

Inter-annotator agreement was much lower: Krippendorff's alpha=0.598. It is because model-generated purpose statements are often ambiguous and perceived differently by annotators.

Examples of the purposes are shown in the Appendix D and typical errors are given in the Appendix E. The errors were related to diary entries:

- mentioning other people's diaries: error in identifying the author (e.g., "to familiarize oneself with Korolenko's true experience");
- mentioning other types of entries: error in identifying the type of entry (e.g., "writing on canvas", "writing down objections to an article");

---

[6] Union with baseline are not given as they show low performance.

- mentioning plans, reasoning or style of keeping a diary, but not the purpose: error in identifying the purpose (e.g. "keep a diary neatly").

**Step 3. Purpose Clustering.** Each LLM clustered all three sets of extracted purposes (separately) according to the proposed algorithm. The number of clusters allocated by each model for each set of purposes is shown in Table 7.

**Table 7.** The number of clusters. The rows correspond to the models that performed clustering of the purpose sets, the columns correspond to the models that generated these purpose sets.

| Clustering model | Purpose extraction model | | |
|---|---|---|---|
| | GPT-4o (141 purposes) | o1-mini (85 purposes) | DeepSeek (221 purposes) |
| GPT-4o | 13 | 9 | 10 |
| o1-mini | 10 | 10 | 10 |
| DeepSeek | 10 | 9 | 12 |
| manual | 11 | 9 | 16 |

We compared the resulting partitions with the manual partitions and calculated the Rand index (Table 8). The Rand index quantifies the similarity between data clusterings by measuring the proportion of pairs of elements that are either assigned to the same cluster or assigned to different clusters. The best results are again GPT-4o and o1-mini, although DeepSeek is not far behind, apart from the case of clustering a large number of purposes (221) extracted by DeepSeek itself. Examples of the clusters are given in the Appendix F.

**Table 8.** Results of the clustering. The rows correspond to the models that performed clustering of the purpose sets, the columns correspond to the models that generated these purpose sets. The values in the cells are the Rand indices ($\uparrow$).

| Clustering model | Purpose extraction model | | |
|---|---|---|---|
| | GPT-4o | o1-mini | DeepSeek |
| GPT-4o | **0.8693** | **0.8333** | 0.8490 |
| o1-mini | 0.8398 | 0.8076 | **0.8537** |
| DeepSeek | 0.8367 | 0.8280 | 0.7956 |

## 6 Discussion

We analyze GPT-4o clusters based on GPT-4o purposes as the best option for clustering the identified purposes (see the first row and first column in Table 8), specifically 13 clusters that contain 126 correctly identified purposes. During this clustering process, the total number of purposes was reduced to 109, as some purposes allocated within a single entry were grouped into the same cluster; such purposes were not considered in further analysis. The name of these clusters and examples of purposes for each cluster are provided in the Appendix F.

Fig. 1 shows the distribution of purposes among the clusters identified by GPT-4o. In the following diagrams, the order of clusters is preserved as in Fig. 1, specifically by their frequency of occurrence.

Three clusters – "Preservation of Memories," "Personal History and Memoirs," and "Memory of Other People" – are related to memories. They differ as follows: "Preservation of Memories" refers to keeping personal memories for oneself. "Personal History and Memoirs" involves preserving personal memories for others. "Memory of Other People" pertains to preserving memories of other individuals. Examples of diary entries from these clusters are shown in Appendix G.

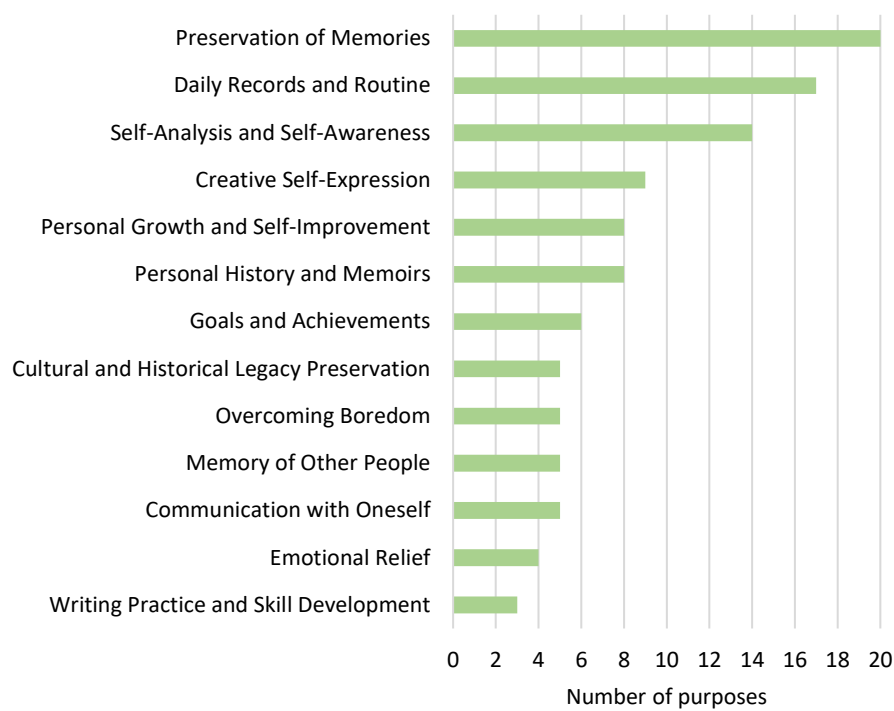

**Fig. 1.** Distribution of purposes among the clusters identified by GPT-4o.

We consider the composition of the clusters by gender, author age category, and period of entries writing (Fig. 2–4).

The original corpus contained 5.4 times more male diary entries than female diary entries (Table 1). The manually annotated part of the corpus (460 entries) consists of 284 male and 176 female entries. When examining entries that specified a purpose, the ratio shifted to 1.6, still favoring male entries. This indicates that women were more inclined to articulate the purpose behind their diary-keeping.

Due to the imbalance in the original corpus, we consider not the number of male and female entries, but the proportion of these entries with purposes in each cluster relative to the total number of male and female entries with purposes in the labeled part of the corpus. Specifically, out of 109 labeled purposes, there were 67 male and 42 female entries. Among the 20 entries containing purposes, the cluster "Preservation of Memories" included 14 male and 6 female entries. Thus, the proportion of male entries in this cluster was 0.21 (14 out of 67), while the proportion of female entries was 0.14 (6 out of 42). Similar calculations were performed for all clusters, and the results are shown in Fig. 2.

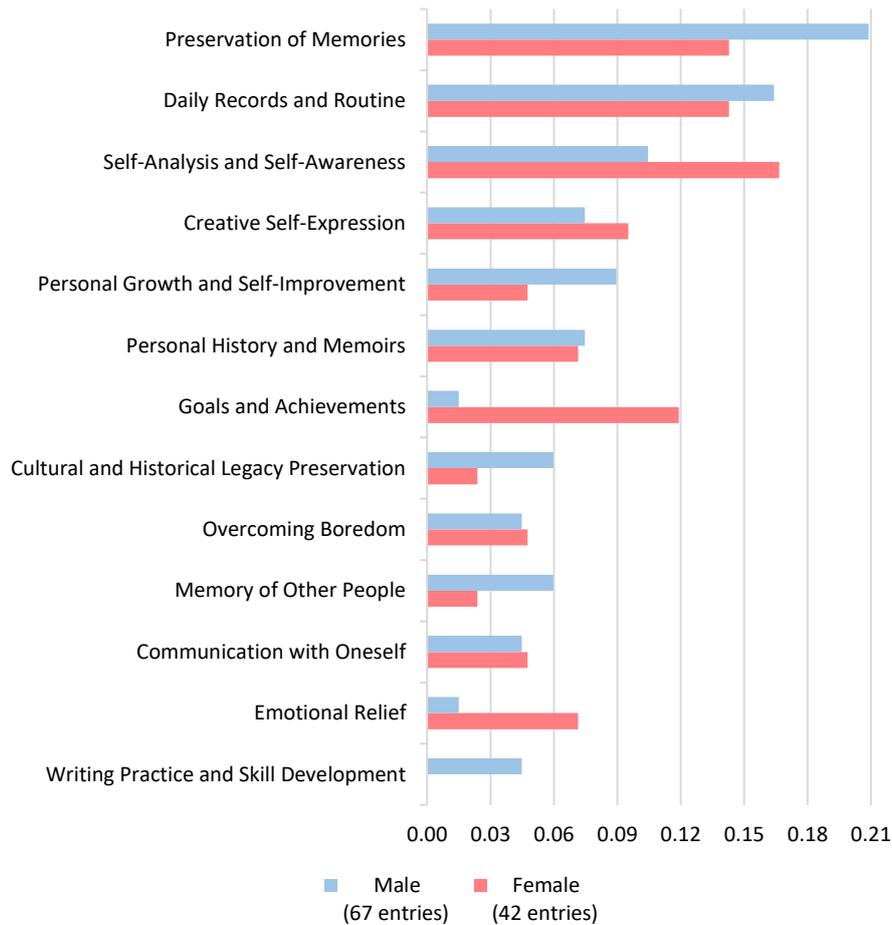

**Fig. 2.** The proportions of male and female entries with purposes in clusters. After the #, the chart legend indicates the number of purposes for each category.

The proportions in Fig. 2 allow us to hypothesize that the purposes of "Personal History and Memoirs" holds equal significance for both genders. Men tend to focus more on retaining memories, personal growth and recording daily events, while women are more likely to emphasize emotional release, achievement and self-analysis in their diary entries.

The analysis of purposes by the author's age category is shown in Fig. 3 (we also use proportions rather than absolute numbers).

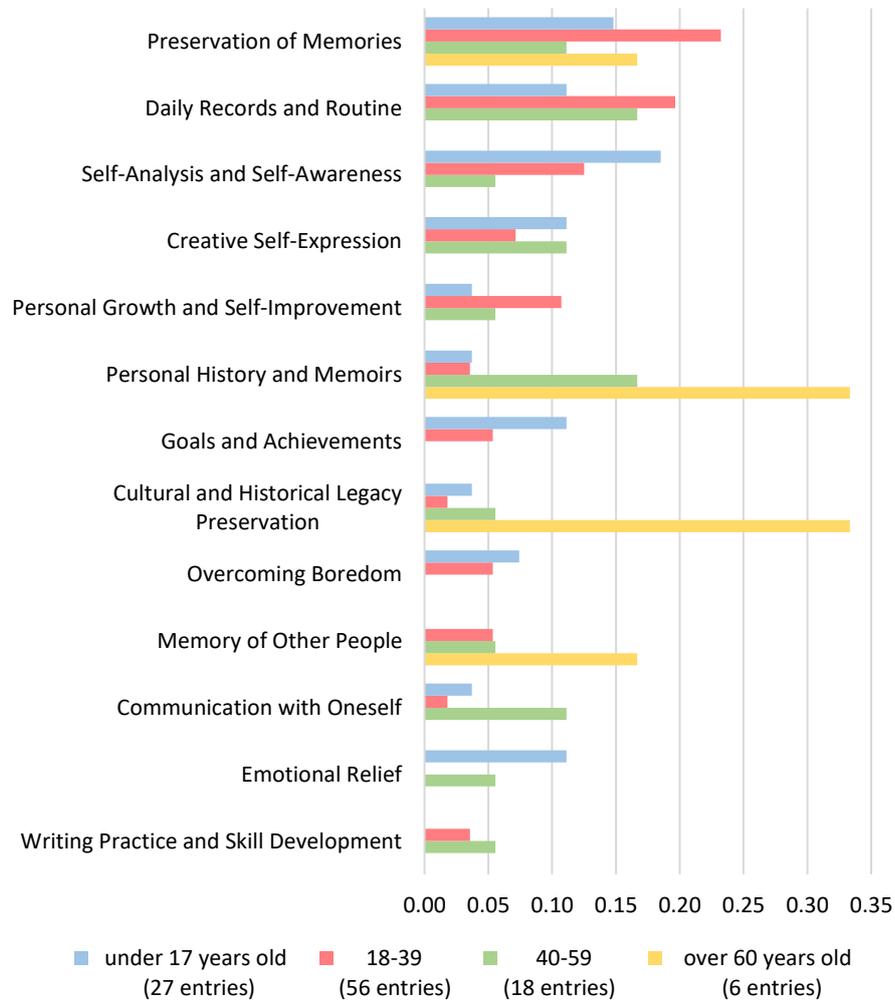

**Fig. 3.** The proportions of age categories in clusters. After the #, the chart legend indicates the number of purposes for each category.

It is worth noting that in the diaries of the age category over 60, only purposes about preserving memories were identified. The under-17 age category focuses on entries related to self-analysis. Interestingly, the cluster "Goals and achievements" and "Overcoming Boredom" is divided among two young age categories (under 17 and 18-39).

The analysis of purposes by the period of writing the entries is shown in Fig. 4.

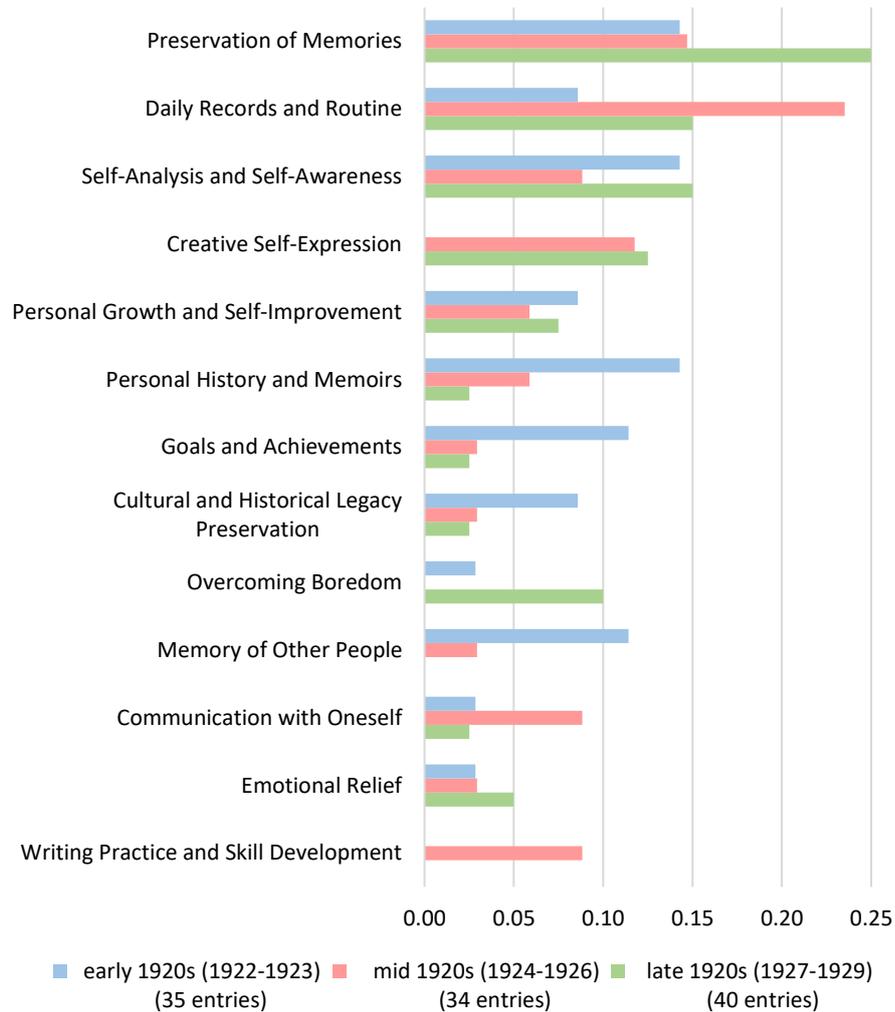

**Fig. 4.** The proportions of periods in clusters. After the #, the chart legend indicates the number of purposes for each category.

The cluster "Preservation of Memories" remains consistently popular across all three periods. Meanwhile, the cluster "Self-Analysis and Self-Awareness" experiences a decline during the mid-1920s. In contrast, the cluster "Daily Records and Routine" sees an increase in frequency during the same timeframe. The "Memory of Other People" decreases from 11% to 0% over three periods, while the "Overcoming Boredom" cluster increases in the later period.

# 7     Conclusion

In this study, we propose a method for extracting and clustering purposes in diary entries using LLMs. We evaluated three state-of-the-art LLMs alongside a template-based baseline. GPT-4o and o1-mini demonstrated the best performance, while the baseline performed poorly.

Our experimental results suggest the following recommendations. If precision is the primary concern, GPT-4o is the best choice. However, if the purpose is to retrieve the maximum number of potentially relevant entries at a lower cost, followed by manual labeling, DeepSeek is preferable. A balanced approach is to combine the results of GPT-4o and o1-mini, offering a trade-off between precision and recall.

We also highlight common model errors found in the extraction of purposes from diary entries, such as misidentifying the author by referencing other people's diaries, confusing diary entries with other types of writing, and mentioning plans or styles of diary-keeping without clearly stating the actual purpose.

The use of LLMs allows to save a lot of time while obtaining acceptable quality of the analysis. According to our experts' estimates, the time one person would spend reading the entire corpus is on the order of 300 hours (37.5 working days), while LLM processed in an hour.

Our analysis of diary-keeping purposes provides insights into the motivations behind personal writing across different demographics. We found that both genders appreciate "Personal History and Memoirs", but men are more focused on memory retention and personal growth, while women place greater emphasis on emotional release and self-analysis.

Age also plays a significant role; individuals over 60 primarily write to preserve memories, whereas those under 17 engage more in self-reflection.

Notably, the "Preservation of Memories" cluster remains consistently popular across all three examined time periods (1922–1923, 1924–1926, 1927–1929). In contrast, "Self-Analysis and Self-Awareness" experiences a decline in the mid-1920s, while the frequency of "Daily Records and Routine" increases during the same timeframe.

## Limitation

It should be noted that the study encounters certain limitations. The main difficulty lies in the ambiguous definition of the concept of "purpose of keeping a diary". This led to the need to write a more detailed prompt and provide examples. This ambiguity was also the main reason for the disagreement between the annotators.

Another limitation was the chosen time period for the diary entries – the 1920s. Misunderstanding of the historical context could lead to hallucination of the models and incorrect labeling by the annotators.

The reasons mentioned above, as well as the small number of annotators (three persons), could lead to bias in the quality assessments of the considered models.

Another issue is that we lack information about the complete set of diary entries in our corpus that contain purposes, as labeling the entire corpus is too time-consuming.

Therefore, we utilize relative Recall (and consequently relative F1-score) since we only consider the set of entries extracted by all four models.

## Ethics Statement

This study was conducted with careful consideration of ethical principles. A diary entry may contain sensitive information that the author may not have intended to share. In our work, we relied on the principles of the "Prozhito" diary corpus[7]. In particular, diaries were not used unless the author or their heirs provided permission for publication. The diary entries were also anonymized: neither the models nor the annotators had information about the author of the diary.

The diary corpus was compiled by both professional publishers and interested volunteers. The original diary entry could be edited by the author, an editor during publication, or a center staff member at the center for annotation. Each participant in the process could have contributed their own personal beliefs and biases, which could have distorted the research results. Like any other personal text, a diary requires critical interpretation from the reader – it is essential to understand that the author may have various motivations for keeping a diary, including the desire to distort or alter the description of events.

The authors were in close contact with the "Prozhito" Center for the Study of Ego-Documents of the European University at St. Petersburg, and regular consultations were held.

---

[7] https://prozhito.org/page/corpus/

## Appendix A. Prompt for purpose extraction

Prompt for purpose extraction is available at the repository, Appendix A.

## Appendix B. Prompt for generating cluster names

Prompt for generating cluster names is available at the repository, Appendix B.

## Appendix C. Prompt for purpose clustering

Prompt for purpose clustering is available at the repository, Appendix C.

## Appendix D. Examples of purposes

Examples of purposes is available at the repository, Appendix D.

## Appendix E. Errors in extracting diary-keeping purposes by LLMs

Errors in extracting diary-keeping purposes by LLMs is available at the repository, Appendix E.

## Appendix F. Examples of clusters

Examples of clusters is available at the repository, Appendix F.

## Appendix G. Examples of diary entries

Examples of diary entries is available at the repository, Appendix G.